\documentclass[11pt]{article}

\usepackage[final]{acl}

\usepackage{times}
\usepackage{latexsym}
\usepackage{url}

\usepackage{algorithm}
\usepackage{algpseudocode}
\usepackage{amsmath}
\usepackage{amssymb}
\usepackage{graphicx}
\usepackage{booktabs}      
\usepackage{multirow}      
\usepackage[table]{xcolor} 
\usepackage{colortbl}      

\usepackage[most]{tcolorbox} 

\usepackage{tcolorbox}
\tcbuselibrary{skins,breakable}
\usepackage{dblfloatfix}   

\usepackage{listings}
\usepackage{xcolor}

\usepackage{caption}

\usepackage[T1]{fontenc}

\usepackage[utf8]{inputenc}

\usepackage{microtype}
\usepackage{enumitem}

\usepackage{inconsolata}

\usepackage{graphicx}
\usepackage{subcaption}

\usepackage{tcolorbox}
\tcbuselibrary{breakable}
\usepackage{xcolor}

\newtcolorbox{PromptCard}[1]{
  enhanced,
  colback=white,
  colframe=black!55,
  boxrule=0.8pt,
  arc=2mm,
  left=4mm, right=4mm, top=3mm, bottom=3mm,
  title={#1},
  colbacktitle=black!78,   
  coltitle=white,
  fonttitle=\bfseries\large,
  attach boxed title to top left={xshift=2mm,yshift=-1.2mm},
  boxed title style={
    sharp corners,
    boxrule=0pt,
    arc=1.6mm,
    left=2.8mm, right=2.8mm, top=1.2mm, bottom=1.2mm
  }
}

\lstdefinestyle{promptstyle}{
  basicstyle=\ttfamily\scriptsize,
  columns=fullflexible,
  keepspaces=true,
  breaklines=true,
  breakatwhitespace=false,
  frame=single,
  framerule=0.3pt,
  rulecolor=\color{black!35},
  xleftmargin=0.3em,
  xrightmargin=0.3em,
  aboveskip=0.6em,
  belowskip=0.4em,
  showstringspaces=false,
  upquote=true
}
\title{Robust Tool Use via \textsc{Fission-GRPO}: Learning to Recover from Execution Errors}

\author{
 \textbf{Zhiwei Zhang\textsuperscript{1,3}},
 \textbf{Fei Zhao\textsuperscript{2}},
 \textbf{Rui Wang\textsuperscript{1,3}},
 \textbf{Zezhong WANG\textsuperscript{1}},
\\
 \textbf{Bin Liang\textsuperscript{1,3}},
 \textbf{Jiakang Wang\textsuperscript{2}},
 \textbf{Yao Hu\textsuperscript{2}},
 \textbf{Shaosheng Cao\textsuperscript{2}}\thanks{\hspace{2pt}Corresponding author.},
 \textbf{Kam-Fai Wong\textsuperscript{1,3}}\footnotemark[\value{footnote}]
\\
\\
 \textsuperscript{1}The Chinese University of Hong Kong,\\
 \textsuperscript{2}Xiaohongshu Inc.,\\
 \textsuperscript{3}MoE Key Laboratory of High Confidence Software Technologies\\
 \texttt{zhangzhiwei1019@link.cuhk.edu.hk, caoshaosheng@xiaohongshu.com}
 }

\begin{document}
\maketitle





\begin{abstract}

Large language models (LLMs) can call tools effectively, yet they remain brittle in multi-turn execution: after a tool-call error, smaller models often fall into repetitive invalid re-invocations instead of interpreting the feedback and recovering.
This failure mode persists because current training paradigms do not explicitly teach models how to recover from execution errors.
In particular, standard reinforcement learning (RL) collapses rich failure experience into sparse negative rewards, while pre-collected error-correction datasets become mismatched to the policy's evolving failure modes.
To bridge this gap, we propose \textsc{Fission-GRPO}, a framework that converts execution errors into on-policy corrective supervision within the RL training loop.
Our core mechanism \textit{fissions} each failed trajectory into a new training instance by augmenting it with diagnostic feedback from a fine-tuned Error Simulator, then resampling multiple recovery rollouts on-policy.
This enables the model to learn from the precise errors it makes during exploration, rather than from static, pre-collected error cases.
On BFCL v4 Multi-Turn, \textsc{Fission-GRPO} improves the error recovery rate of Qwen3-8B by 5.7\% absolute and overall accuracy by 4.0\% (from 42.75\% to 46.75\%), outperforming both RL baselines and specialized tool-use agents. The method further generalizes to TAU-Bench and TAU2-Bench, achieving leading results across most settings with gains up to +17.4\%.\footnote{Our code is available at \url{https://github.com/zxzadm/Fission-GRPO}.}

\end{abstract}

 \section{Introduction}

\begin{figure}[t!]
  \centering

  \begin{subfigure}[t]{0.9\columnwidth}
    \centering
    \includegraphics[width=\linewidth]{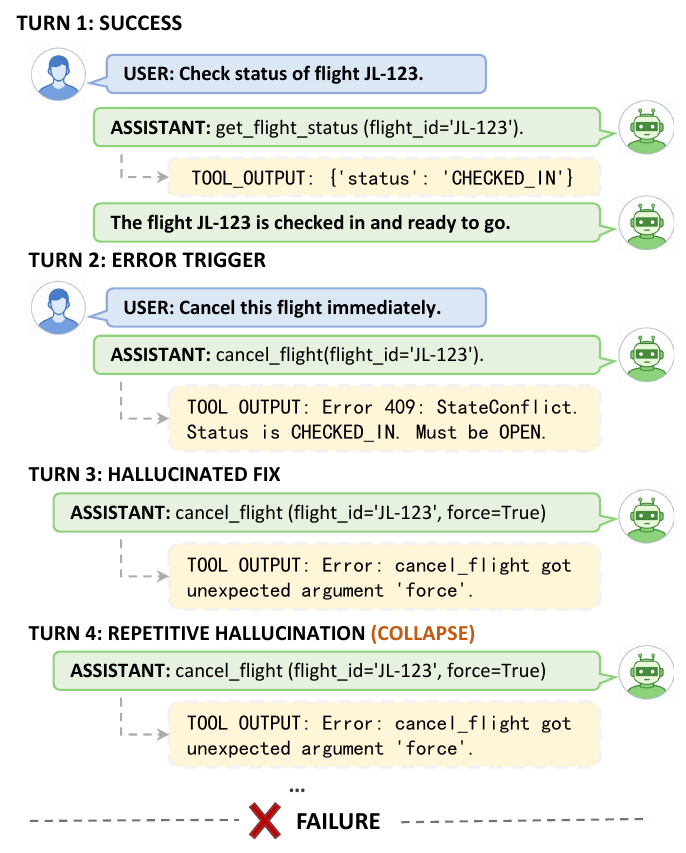}
    \caption{Typical failure: an API error triggers a hallucinated retry loop.}
    \label{fig:failure_pattern}
  \end{subfigure}

  \vspace{0.4em}

  \begin{subfigure}[t]{0.9\columnwidth}
    \centering
    \includegraphics[width=\linewidth]{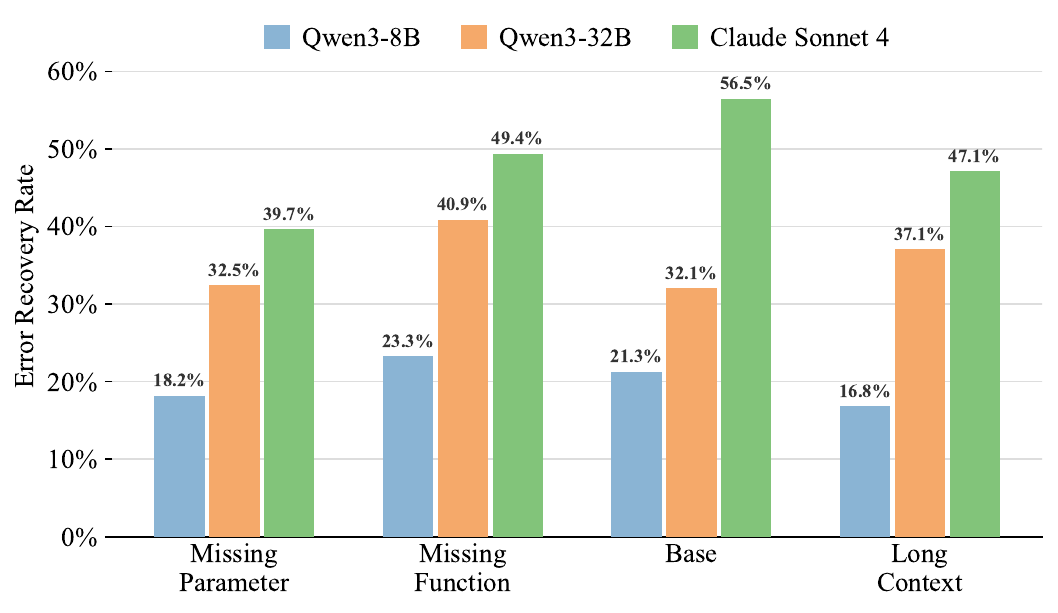}
    \caption{Error recovery rates on BFCL v4 Multi-Turn across model scales and evaluation subsets.}
    \label{fig:recovery_rates}
  \end{subfigure}

  \caption{Error recovery is a key bottleneck for smaller tool-using models in multi-turn execution.
  (a) shows a representative hallucinated retry loop after an API error, while (b) reports recovery rates on BFCL v4 across model scales.}
  \label{fig:error_recovery_challenge}
  \vspace{-4mm}
\end{figure}

Tool-using agents are increasingly moving beyond static text prediction toward interactive decision-making grounded in environment feedback.
A growing body of work argues that the next wave of capability gains will come not from scaling human-curated data alone, but from \textit{experience}---data generated by agents interacting with their environments~\cite{silver2025welcome}.
Multi-turn tool use is already an instance of this transition: the agent acts via tool calls, observes execution results, and must adapt to changing state and error signals across turns.
For such agents to be deployed reliably, especially at smaller scales suitable for low-latency and on-device settings~\cite{belcak2025small, sharma2025small}, they must exhibit \textit{robustness}---the ability to recover from the execution errors that inevitably arise in dynamic, multi-turn environments~\cite{patilberkeley}.

This robustness requirement exposes a critical gap.
In practice, APIs return errors, parameters become invalid, and system states change unexpectedly; a robust agent must interpret such feedback, diagnose the fault, and self-correct~\cite{yao2022react, shinn2023reflexion}.
Yet as shown in Figure~\ref{fig:recovery_rates}, smaller models exhibit a pronounced deficiency in \textit{error recovery}---the probability of eventual success conditioned on at least one prior execution error.
On BFCL v4 Multi-Turn~\cite{patilberkeley}, Claude Sonnet 4 exceeds 50\% recovery while Qwen3-8B averages only around 20\%.
Figure~\ref{fig:failure_pattern} illustrates the typical failure mode---hallucinated retries that loop until the conversation collapses.

Current approaches fall short of addressing this challenge.
Methods based on \textbf{static synthetic datasets}~\cite{liu2024toolace, zhang2025xlam, zhang2025looptool} construct error-correction pairs offline, but the error distribution shifts as the policy improves, making offline error corpora quickly stale---a manifestation of the broader limitation that static, pre-collected data cannot keep pace with an evolving agent~\cite{silver2025welcome}.
Meanwhile, \textbf{reinforcement learning (RL)} approaches such as GRPO~\cite{shao2024deepseekmath} treat errors merely as sparse negative rewards.
This signals that something went wrong, but offers no guidance on \textit{how} to recover: the gradient discourages the failed action without teaching a corrective alternative.
When all sampled rollouts fail, the advantage variance collapses, yielding vanishing gradients that stall learning entirely~\cite{yu2025dapo, nan2025ngrpo}.
In essence, existing methods treat errors as outcomes to be \textit{avoided} rather than experiences to be \textit{learned from}.

\begin{figure*}[t]
    \centering
    \includegraphics[width=\textwidth]{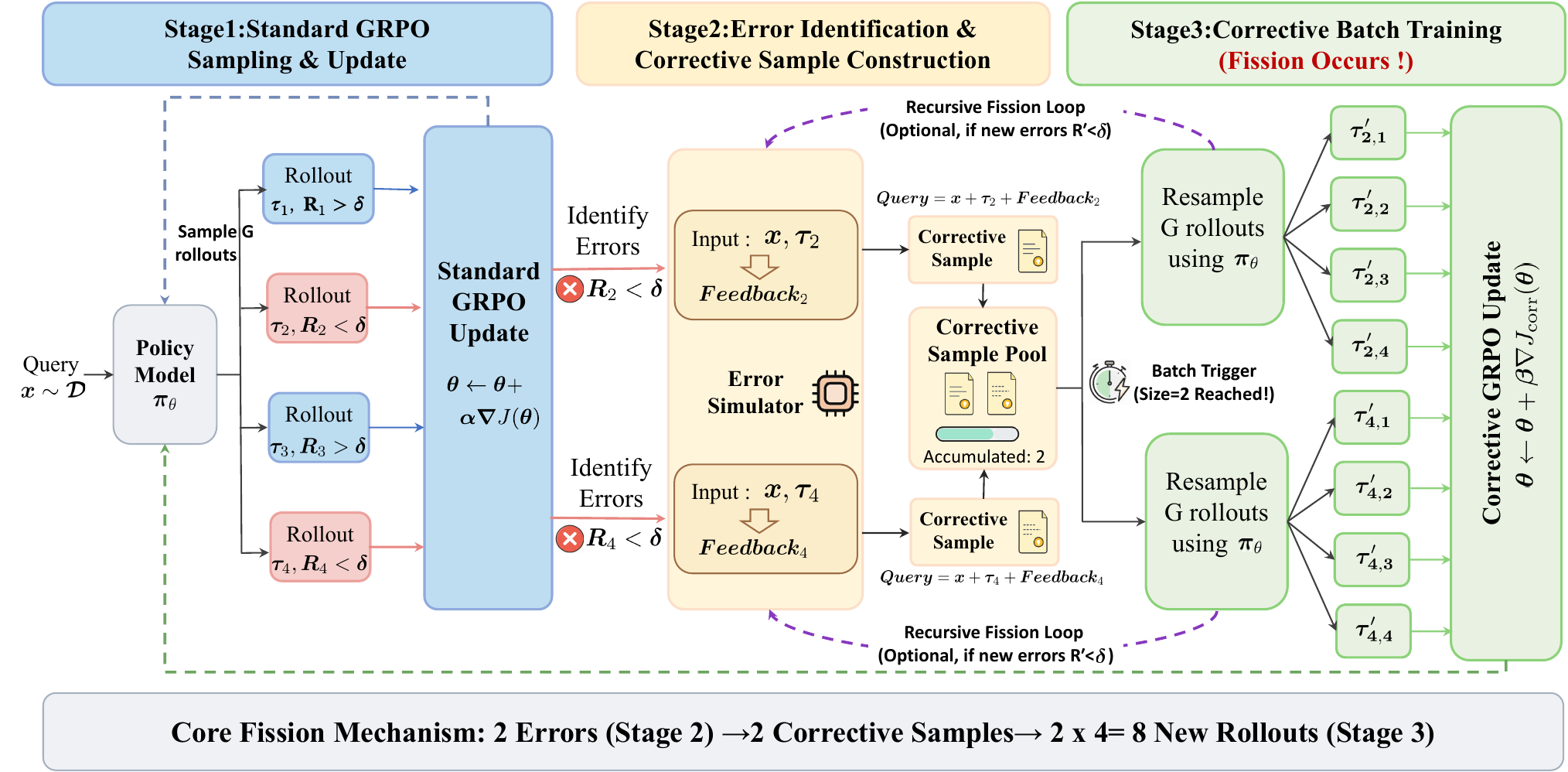} 
    \caption{\textbf{Overview of the \textsc{Fission-GRPO} Framework.} The framework operates in three stages: (1) \textbf{Standard Exploration}, utilizing GRPO to optimize policy $\pi_\theta$ on the query distribution $\mathcal{D}$; (2) \textbf{Error Identification \& Synthesis}, where a simulator $\mathcal{S}_\phi$ generates diagnostic feedback for filtered error trajectories; and (3) \textbf{Fission-based Update}, where corrective samples trigger a multiplicative resampling process (factor $G'$) to align the policy with recovery paths.}
    \label{fig:method_overview}
\end{figure*}

To bridge this gap, we propose \textsc{Fission-GRPO}, a framework that transforms execution errors into dense, on-policy corrective experience (Figure~\ref{fig:method_overview}).  
In \textbf{Stage 1}, we perform standard GRPO exploration, sampling multiple rollouts per query and updating the policy with group-relative advantages.  
In \textbf{Stage 2}, failed rollouts are intercepted and augmented with diagnostic feedback from a learned Error Simulator, constructing corrective contexts of the form [\textit{dialogue}; \textit{failed call}; \textit{feedback}].  
In \textbf{Stage 3}, these contexts trigger a \textit{fission} update: each error is expanded into $G'$ parallel recovery attempts by resampling new rollouts from the augmented context---analogous to nuclear fission, where a single event induces a chain of reactions, generating dense training signals from a single failure.

The Error Simulator is trained via supervised fine-tuning to produce realistic, context-aware diagnostics resembling runtime error traces, with outputs restricted to non-revealing descriptions (e.g., ``parameter \texttt{status} expects value \texttt{OPEN}'') to prevent target leakage.
This closed-loop process continuously focuses learning on the model's current error modes, mitigating the distribution mismatch inherent in static error-correction datasets.

We evaluate \textsc{Fission-GRPO} on BFCL v4 Multi-Turn, TAU-Bench~\cite{yao2024tau}, and TAU2-Bench~\cite{barres2025tau}, with consistent improvements across benchmarks and model scales.
Our main contributions are:

\begin{itemize}[leftmargin=*, noitemsep, topsep=0pt]
  \item \textbf{Fission-GRPO Framework.} We propose an RL framework that dynamically converts execution errors into corrective training instances. By resampling from augmented error contexts on-policy, our approach maintains alignment with the model's evolving error distribution.
  
  \item \textbf{Learned Error Simulator.} We develop a fine-tuned error simulator that generates realistic diagnostic feedback without target leakage, enabling effective recovery training without live API interactions. Cross-domain evaluations on unseen tool schemas and human evaluation (96\% non-leakage; Cohen's $\kappa = 0.71$) confirm its reliability and generalizability.
  
  \item \textbf{Empirical Validation.} \textsc{Fission-GRPO} achieves state-of-the-art performance on BFCL v4 Multi-Turn across the Qwen3 model family (1.7B, 4B, and 8B). For Qwen3-8B, it improves error recovery by 5.7\% absolute and overall accuracy by 4.0\% (42.75\%\,$\rightarrow$\,46.75\%). The method further generalizes to TAU-Bench and TAU2-Bench with entirely different tool ecosystems, achieving leading results across most settings with gains up to +17.4\% on TAU1 Retail.
\end{itemize}

\section{Related Work}

\subsection{RL for Tool Use}
RL has become the standard for aligning LLMs~\cite{schulman2017proximal, ouyang2022training}. Among recent algorithms, GRPO~\cite{shao2024deepseekmath} eliminates the value network by estimating baselines from group averages, making it well-suited to tool-calling tasks with binary or scalar rewards~\cite{guo2025deepseek}.

However, GRPO's reliance on intra-group variance creates a failure mode: when a sampled group is homogeneously incorrect, reward variance vanishes and gradients become null---a limitation targeted by DAPO~\cite{yu2025dapo} and NGRPO~\cite{nan2025ngrpo}. Even when gradients exist, indiscriminate negative feedback can trigger \textit{Lazy Likelihood Displacement (LLD)}~\cite{deng2025effect}, suppressing valid reasoning steps merely because they co-occur with failed trajectories.

Existing mitigations---filtering homogeneous batches (DAPO), calibrating advantages (NGRPO), or down-weighting negative gradients (NTHR~\cite{deng2025effect})---reshape the \textit{loss landscape} of negative signals but leave the scarcity of positive guidance during exploration unaddressed. Our approach instead actively constructs recovery trajectories via fission, turning zero-reward errors into dense learning signals.

\subsection{Robust Tool Use and Error-Driven Synthesis}
Research in tool utilization has evolved from single-turn syntactic correctness~\cite{schick2023toolformer, patil2024gorilla} to reliability across multi-turn workflows~\cite{qin2023toolllm, yao2022react} and human-level task solving~\cite{webaggregator,cogkernal}. As tasks grow more complex, recovery from inevitable environment errors (e.g., timeouts, invalid parameters) becomes a defining metric of robustness, as codified in benchmarks like BFCL~\cite{patilberkeley} and StableToolBench~\cite{guo2024stabletoolbench}.

Recent work addresses this via ``diagnosis-and-repair'' mechanisms~\cite{su2025failure, huang2025critictool} or training on diverse error scenarios~\cite{vuddanti2025paladin}. Synthetic correction methods, proven in reasoning domains~\cite{pan2025lemma, xu2025subtle}, have been adapted to tool use by frameworks like ToolACE~\cite{liu2024toolace} and LoopTool~\cite{zhang2025looptool} to expand training coverage through model-based synthesis.

Yet these methods rely predominantly on \textit{offline} data construction, creating a temporal mismatch with the policy's evolving on-policy error distribution~\cite{kumar2024training, zhang2025looptool,fang2025webevolver,li2026cso}. Unlike prior offline synthesis~\cite{pan2025lemma, su2025failure}, our work integrates error simulation directly into the training loop, keeping supervision aligned with current policy limitations.

\section{Method}
\label{sec:method}

We propose \textsc{Fission-GRPO}, a framework designed to imbue small language models with robust error recovery capabilities. As illustrated in Figure~\ref{fig:method_overview}, our approach operates in a three-stage closed loop: standard GRPO exploration maintains general tool-use competence, while a conditional fission stream intercepts errors and triggers active remedial learning.

\subsection{Preliminaries}
\label{sec:prelim}
We formulate tool use as a language generation task. Given a query $x$ and a tool library, a policy $\pi_\theta$ generates a trajectory $\tau$ consisting of reasoning thoughts and tool calls.
We adopt \textbf{GRPO}~\cite{shao2024deepseekmath} as our optimization backbone. Unlike PPO, GRPO eliminates the need for a value network by estimating the baseline from the group average. For each query $x$, we sample a group of outputs $\{\tau_i\}_{i=1}^G$ and optimize:
\begin{equation}
\label{eq:grpo_obj}
\mathcal{J}(\theta) = \mathbb{E}_{x \sim \mathcal{D}} \left[ \frac{1}{G} \sum_{i=1}^G \hat{A}(\tau_i) \cdot \pi_{\text{ratio}}(\tau_i \mid x) - \beta \mathbb{D}_{\text{KL}} \right]
\end{equation}
where $\hat{A}(\tau_i) = \frac{R(\tau_i) - \mu_R}{\sigma_R + \epsilon}$ is the group-normalized advantage, with $\mu_R$ and $\sigma_R$ being the mean and standard deviation of rewards within the group, and $\pi_{\text{ratio}}(\tau_i \mid x)$ is the clipped probability ratio.

\subsection{Reward Design}
\label{sec:reward}
To guide the policy from syntactic compliance to semantic precision,
we design a time-dependent composite reward with normalized aggregation.
Each component is scaled by a time-varying weight, and the final reward
is the normalized sum:
\begin{equation}
\label{eq:reward}
\begin{aligned}
R(\tau, t) = \frac{1}{3}\Big[\,
  & w_{\text{fmt}}(t)\, R_{\text{fmt}}(\tau) \\
  & + w_{\text{corr}}(t)\, R_{\text{corr}}(\tau)
    + R_{\text{len}}(\tau) \,\Big]
\end{aligned}
\end{equation}
where each weighted term is bounded such that $R(\tau, t) \in [0, 2]$
throughout training.

\paragraph{Format Compliance ($R_{\text{fmt}}$).}
This binary term $R_{\text{fmt}}(\tau) \in \{0, 1\}$ enforces structural
constraints, ensuring outputs adhere to the required XML/JSON schema.
We apply a decaying weight $w_{\text{fmt}}(t)$ that anneals the maximum
weighted contribution from 3 to 2, shifting focus from syntax to
semantics as training progresses.

\paragraph{Functional Correctness ($R_{\text{corr}}$).}
This term evaluates alignment between invoked tools and user intent.
To accommodate partial matching in complex parameters, we define
$R_{\text{corr}}(\tau, y^*) \in [0, 1]$ as:
\begin{equation}
\begin{aligned}
R_{\text{corr}}(\tau, y^*) = {}& \alpha \cdot \mathbb{I}(N = N^*) \\
&+ (1-\alpha) \cdot \frac{1}{|\mathcal{M}|}\sum_{(a, a^*) \in \mathcal{M}} \text{F1}(a, a^*)
\end{aligned}
\end{equation}
where $\alpha \in [0,1]$ (set to $\alpha=0.5$ in our experiments)
balances function selection against parameter matching, $\mathbb{I}(N=N^*)$
indicates correct function selection, $\mathcal{M}$ denotes matched
argument pairs, and F1 measures token-level overlap. The weight
$w_{\text{corr}}(t)$ increases monotonically, scaling its maximum
weighted contribution from 2 to 3 to prioritize parameter precision
in later stages.


\paragraph{Efficiency Regularization ($R_{\text{len}}$).}
To prevent verbose or degenerate reasoning, we add a length-compliance reward $R_{\text{len}} \in [0, 1]$, computed via a piecewise Gaussian function that peaks at the target length and decays on both sides.

\subsection{The \textsc{Fission-GRPO} Framework}
\label{sec:framework}

As illustrated in Figure~\ref{fig:method_overview}, \textsc{Fission-GRPO} operates in a three-stage closed loop. Stage 1 focuses on optimizing \textbf{fundamental tool-use capabilities}, while Stages 2 and 3 are dedicated to developing \textbf{error recovery skills} through targeted error correction.

\subsubsection{Stage 1: Standard Exploration and Update}
\label{sec:stage1}
This stage aims to establish and maintain the model's base performance on tool-calling tasks.

\paragraph{Sampling and Evaluation.} 
Given a query $x$, we sample a group of trajectories $\{\tau_i\}_{i=1}^G$ from the current policy $\pi_\theta$. We evaluate these rollouts using the composite reward function defined in \S\ref{sec:reward}, computing format compliance $R_{\text{fmt}}$, functional correctness $R_{\text{corr}}$, and efficiency regularization $R_{\text{len}}$, which are then aggregated into the total reward.

\paragraph{Optimization.} 
We apply the standard GRPO update (Eq.~\ref{eq:grpo_obj}) using these trajectories to improve the model's fundamental tool-use capabilities. All sampled trajectories are then forwarded to Stage 2 for diagnostic error analysis and corrective training.

\subsubsection{Stage 2: Error Identification and Corrective Sample Construction}
\label{sec:stage2}

Stage~2 converts error traces produced in Stage~1 into actionable corrective instances. Concretely, we apply a two-level filter to isolate erroneous trajectories and then synthesize feedback that can be appended to the original context for subsequent corrective updates.

\paragraph{Error Identification.}
We decompose error detection into \emph{format validity} and \emph{functional correctness}. Let $R_{\text{fmt}}$ denote whether the tool-call format is valid. If $R_{\text{fmt}}(\tau)=0$, the trajectory is immediately treated as an error without consulting correctness. Otherwise, we further evaluate correctness with a scalar score $R_{\text{corr}}$ and flag the trajectory when it falls below a tunable threshold $\delta_{\text{corr}}$:
\begin{equation}
\label{eq:failset}
\mathcal{E} = \{ \tau_i \mid R_{\text{corr}}(\tau_i) < \delta_{\text{corr}} \lor R_{\text{fmt}}(\tau_i) = 0 \}
\end{equation}
In Fig.~2, we use a simplified illustration (e.g., $R<\delta$) to emphasize the gating effect; this does not contradict Eq.~\eqref{eq:failset}.

\paragraph{Hybrid Feedback Synthesis.}
For effective correction, a scalar penalty is insufficient; we require an explicit diagnostic message $f$ that resembles the runtime system feedback. We adopt a hybrid strategy:
(i) for \emph{format errors} ($R_{\text{fmt}}=0$), we use deterministic error messages (e.g., parser/compiler-style feedback) to explicitly state the violated schema/serialization constraints; 
(ii) for \emph{semantic errors} ($R_{\text{corr}}<\delta_{\text{corr}}$), we query a learned \textbf{Error Simulator} $S_{\phi}$ to produce a concise, actionable runtime error string.

The simulator is implemented as a Qwen3-32B model fine-tuned via SFT to emulate runtime environment responses. We construct a training set of approximately 2K instances from error logs, where each instance comprises: (i) the original system prompt and tool specification along with the dialogue state, (ii) the model's failed tool call ($\tau_{\text{err}}$), (iii) the ground-truth tool call ($\tau_{\text{gt}}$), and (iv) a teacher-written diagnostic error message (via Claude Sonnet 4), after quality filtering. During both training and inference, the simulator consumes $(\text{system\,+\,tools},\ \text{dialogue history},\ \tau_{\text{gt}},\ \tau_{\text{err}})$ and produces a concise feedback string $f \leftarrow S_{\phi}(x,\tau_{\text{err}},\tau_{\text{gt}})$,
where $f$ is constrained to be a realistic runtime response. We provide the exact prompting template used to query $S_{\phi}$ in Appendix~\ref{app:simulator_prompt}.

\paragraph{Corrective Sample Construction and LIFO Buffering.}
Given a flagged trajectory $\tau_{\text{err}}$ with feedback $f$, we construct a corrective context $x_{\text{corr}} = [x; \tau_{\text{err}}; f]$ by appending the failed attempt and the diagnostic message to the original multi-turn input.
We optionally deduplicate corrective instances by hashing the pair $(x, \tau_{\text{err}})$, so that distinct diagnostic messages synthesized for the same underlying error are not treated as separate training instances.
All corrective samples are stored in a \textbf{LIFO} buffer $\mathcal{B}_{\text{corr}}$, so that the most recent errors are consumed first during corrective updates. This design keeps the corrective batch distribution closer to the current policy $\pi_{\theta}$, improving the on-policy approximation in multi-turn tool-use training.

\subsubsection{Stage 3: Corrective Batch Training}
\label{sec:stage3}

Once the LIFO buffer accumulates sufficient \emph{recent} errors (Batch Trigger), we activate \textbf{Fission} to perform targeted remedial updates for recovery.

\paragraph{Multiplicative Resampling.}
We pop the freshest corrective contexts $x_{\text{corr}}$ and, for each of them, sample a ``fission group'' of $G'$ trajectories conditioned on the same context:
\begin{equation}
    \{\tau'_j\}_{j=1}^{G'} \sim \pi_\theta(\cdot \mid x_{\text{corr}}).
\end{equation}
This turns a single error case into multiple parallel recovery attempts, densifying training signals around the observed error.

\paragraph{More Informative Advantages.}
Hard queries can yield near-homogeneous outcomes in standard exploration, weakening within-group relative advantages. Conditioning on explicit feedback $f$ typically increases outcome diversity within the fission group, improving the usefulness of advantage estimates for recovery updates. For each corrective context $x_{\text{corr}}$, we compute rewards over the sampled recovery trajectories and normalize them within the fission group: $\hat{A}(\tau'_j) = (R(\tau'_j) - \mu_R')/(\sigma_R' + \epsilon)$, where $\mu_R'$ and $\sigma_R'$ denote the mean and standard deviation of rewards within the corrective group.

We then optimize the same GRPO-style clipped surrogate objective as in Eq.~\ref{eq:grpo_obj}, but over the corrective distribution:
\begin{equation}
\label{eq:corr_obj}
\begin{aligned}
\mathcal{J}_{\text{corr}}(\theta) = \mathbb{E}_{x_{\text{corr}}}\Big[ \,
  & \frac{1}{G'}\sum_{j=1}^{G'}
  \hat{A}(\tau'_j)\cdot \pi_{\text{ratio}}(\tau'_j \mid x_{\text{corr}}) \\
  & - \beta \mathbb{D}_{\text{KL}} \, \Big].
\end{aligned}
\end{equation}
This corrective objective preserves the optimization form of standard GRPO while shifting the training distribution toward the policy's current failure modes. As a result, the model is explicitly optimized not only to avoid errors, but also to recover from them under feedback-augmented contexts.

\paragraph{Summary.}
These three stages form a continuous loop; detailed pseudocode and hyperparameters are provided in Algorithm~\ref{alg:fission_grpo} (Appendix~\ref{app:algorithm}).

\section{Experiments}
\label{sec:exp}

\begin{table*}[t]
\centering
\scriptsize
\setlength{\tabcolsep}{4.2pt}
\renewcommand{\arraystretch}{1.15}

\resizebox{\textwidth}{!}{%
\begin{tabular}{@{} l | ccccc | cc | ccc @{}}
\toprule
\multirow{2}{*}{\textbf{Method}}
& \multicolumn{5}{c|}{\textbf{BFCL v4 Multi-Turn}}
& \multicolumn{2}{c|}{\textbf{TAU-Bench}}
& \multicolumn{3}{c}{\textbf{TAU2-Bench}} \\
\cmidrule(lr){2-6} \cmidrule(lr){7-8} \cmidrule(lr){9-11}
& Overall & Base & Miss Func & Miss Param & Long Ctx
& Retail & Airline
& Retail & Airline & Telecom \\
\midrule

\rowcolor{gray!12}
\multicolumn{11}{l}{\textit{\textbf{Qwen3-1.7B Models}}} \\
\midrule
Base          & 7.80  & 10.00 & 11.00          & 8.00           & 2.50
              & 6.1           & 14.0           & 7.9           & 12.0           & 25.0 \\
GRPO          & 17.12 & 22.00 & \textbf{18.50} & 15.50          & 12.50
              & 7.0           & 20.0           & 8.5           & 12.7           & 32.5 \\
DAPO          & 16.00 & 22.00 & 17.00          & 14.00          & 11.00
              & \textbf{8.7}  & 18.0           & 6.7           & 14.0           & 28.3 \\
Dr.GRPO       & 16.12 & 19.50 & 17.50          & 14.50          & 13.00
              & 7.8           & 22.0           & \textbf{9.4}  & 15.3           & 32.5 \\
\rowcolor{blue!12}
\textbf{\textsc{Fission-GRPO}}
              & \textbf{20.38} & \textbf{29.00} & \textbf{18.50} & \textbf{16.00} & \textbf{18.00}
              & 7.8           & \textbf{24.0}  & 8.5           & \textbf{16.0}  & \textbf{37.5} \\
\midrule

\rowcolor{gray!12}
\multicolumn{11}{l}{\textit{\textbf{Qwen3-4B Models}}} \\
\midrule
Base          & 19.37 & 24.50 & 19.00 & 14.50          & 19.50          & 19.1          & 26.0          & 22.8          & 20.0          & 22.5 \\
GRPO          & 36.38 & 46.50 & 34.50 & 27.50          & 37.00          & 20.0          & 24.0          & 27.2          & 26.0          & 37.5 \\
DAPO          & 38.25 & 48.50 & 36.50 & 28.00          & \textbf{40.00} & 21.7          & \textbf{32.0} & 28.1          & 24.0          & 30.0 \\
Dr.GRPO       & 34.75 & 43.00 & 34.50 & 27.50          & 34.00          & 20.9          & 22.0          & 23.7          & 26.0          & 35.0 \\
AWPO          & 38.75 & 43.00 & 39.50 & \textbf{36.50} & 36.00          & 27.0          & 22.0          & \textbf{29.8} & 28.0          & 40.0 \\
\rowcolor{blue!12}
\textbf{\textsc{Fission-GRPO}}
              & \textbf{40.87} & \textbf{51.50} & \textbf{42.50} & 30.50 & 39.00
              & \textbf{37.4}  & 30.0          & 29.0          & \textbf{32.0} & \textbf{42.5} \\
\midrule

\rowcolor{gray!12}
\multicolumn{11}{l}{\textit{\textbf{External 8B Agent Models}}} \\
\midrule
ToolACE-2-8B  & 37.00 & 47.00 & 31.00 & 28.00 & 42.00 & 0.9 & 4.0 & 8.2 & 26.7 & 26.7 \\
BitAgent-8B   & 37.75 & 46.50 & 37.50 & 24.00 & 43.00 & 2.6 & 6.0 & 6.1 & 37.3 & 6.7  \\
\midrule

\rowcolor{gray!12}
\multicolumn{11}{l}{\textit{\textbf{Qwen3-8B Models}}} \\
\midrule
Base          & 28.75 & 35.50 & 37.00          & 22.50          & 20.00          & 35.7          & 23.2          & 29.8          & 28.0          & 40.0 \\
GRPO          & 42.75 & 50.50 & 41.00          & 36.00          & 43.50          & 39.1          & 36.0          & 28.1          & 26.0          & 52.5 \\
DAPO          & 43.12 & 54.50 & 44.50          & 29.00          & 44.50          & 45.2          & 20.0          & 34.2          & 26.0          & 47.5 \\
Dr.GRPO       & 44.88 & 55.00 & \textbf{45.00} & 32.50          & 47.00          & 47.0          & 28.0          & \textbf{39.5} & \textbf{34.0} & 35.0 \\
AWPO          & 44.50 & 53.50 & 41.50          & \textbf{40.50} & 42.50          & 43.5          & 30.0          & 32.5          & 26.0          & 45.0 \\
\rowcolor{blue!12}
\textbf{\textsc{Fission-GRPO}}
              & \textbf{46.75} & \textbf{57.50} & 43.50 & 38.00 & \textbf{48.00}
              & \textbf{51.3}  & \textbf{40.0} & 36.8           & 32.0  & \textbf{55.0} \\
\bottomrule
\end{tabular}%
}
\caption{
Main results on BFCL v4 Multi-Turn, TAU-Bench, and TAU2-Bench.
BFCL reports overall and category-level accuracy (\%), while TAU-Bench and TAU2-Bench report pass@1 (\%) under simulated-user interaction.
The best result within each model-size group is highlighted in \textbf{bold}; rows shaded in blue denote our proposed \textsc{Fission-GRPO}.
}
\label{tab:main_results_unified}
\end{table*}

\subsection{Experimental Setup}

\paragraph{Data Construction}
Diverging from prevalent tool-learning paradigms that emphasize extensive scaling of synthetic corpora (e.g., ToolACE~\cite{liu2024toolace}, XLAM~\cite{zhang2025xlam}), we prioritize \textit{data quality} and \textit{trajectory correctness}. We implement a three-stage pipeline to construct a compact yet rigorous training set:

(1) \textbf{Domain Schema Curation}: We curated a diverse schema library spanning 11 domains (e.g., \textit{Healthcare, Smart Home, Vehicle Control}), prompting Claude Sonnet 4 to generate realistic API definitions grounded in BFCL characteristics.

(2) \textbf{Trajectory Synthesis}: Utilizing Claude Sonnet 4, we first synthesized multi-turn user queries based on these schemas, followed by generating full interaction trajectories that fulfill the requests.

(3) \textbf{Hierarchical Filtering and Factorization}: To ensure rigorous quality control, we applied a hierarchical protocol. First, raw trajectories underwent a global coherence check via Claude Sonnet 4. Validated trajectories of length $K$ were then factorized into discrete decision instances $\{(h_t, a_t)\}_{t=1}^K$, where $h_t$ denotes the cumulative context history. Finally, these decomposed instances underwent a dual-model verification via 
Qwen3-235B-A22B-Instruct-2507~\cite{qwen3technicalreport} and Kimi K2~\cite{team2025kimi}. Only samples achieving unanimous consensus were retained, distilling an initial pool of $\sim$2,000 trajectories down to 630 high-quality training instances.

\paragraph{Training Details.}
All models are trained using the \textbf{Verl} framework~\cite{sheng2024hybridflow} on a single node with 8$\times$H800 80GB GPUs.
For GRPO training, we use a learning rate of 1e-6 with cosine warmup, a batch size of 8, and sample 8 rollouts per query ($G=8$).
The maximum prompt length is set to 12,800 tokens and the maximum response length to 4,096 tokens.
We use temperature 0.95 and top-$k$ 50 for sampling.
For \textsc{Fission-GRPO}, we set the correctness threshold $\delta_{\text{corr}}=1$ for error identification (Eq.~\ref{eq:failset}), determined empirically as a stable threshold across multiple runs.

\paragraph{Benchmarks.}
We evaluate on three complementary multi-turn tool-use benchmarks, both of which feature interactive error feedback mechanisms that permit the agent to retry after execution errors---directly aligning with our focus on error recovery.
\textbf{BFCL v4 Multi-Turn}~\cite{patilberkeley} stress-tests state tracking and robustness across up to 20 retry attempts per error, enabling fine-grained measurement of how error recovery dynamics translate to overall success.
\textbf{TAU-Bench}~\cite{yao2024tau} and \textbf{TAU2-Bench}~\cite{barres2025tau} feature genuine multi-turn interactions with LLM-simulated users, employing tool APIs, conversation dynamics, and error distributions substantially different from BFCL. We use GLM-5 as the user simulator and report the pass@1 metric across five settings (Retail, Airline for TAU1; Retail, Airline, Telecom for TAU2).

\paragraph{Baselines.}
We compare \textsc{Fission-GRPO} against RL baselines implemented on the Qwen3 series (1.7B/4B/8B), including:
(1) \textbf{GRPO}~\cite{shao2024deepseekmath}, utilizing group-normalized advantages;
(2) \textbf{DAPO}~\cite{yu2025dapo}, incorporating dynamic sampling constraints;
(3) \textbf{Dr.GRPO}~\cite{liu2025understanding}, employing mean-centered estimators to mitigate length bias; and
(4) \textbf{AWPO}~\cite{lin2025awpo}, which introduces adaptive reward weighting to improve GRPO for tool-use tasks.
For broader context, we also report performance of specialized 8B-scale tool agents such as ToolACE~\cite{liu2024toolace} and BitAgent.

\subsection{Main Results}
\label{sec:main_results}
Table~\ref{tab:main_results_unified} presents results on BFCL v4 Multi-Turn, TAU-Bench, and TAU2-Bench. \textsc{Fission-GRPO} delivers strong and consistent gains across all Qwen3 scales (1.7B, 4B, and 8B) relative to standard GRPO and other post-training baselines. On Qwen3-1.7B, \textsc{Fission-GRPO} lifts BFCL accuracy from 7.80\% to \textbf{20.38\%} 
(an absolute gain of 12.58 points). The advantage persists at larger scales: Qwen3-4B and Qwen3-8B reach \textbf{40.87\%} and \textbf{46.75\%} BFCL accuracy, while the 8B model further achieves the top TAU-Bench scores of \textbf{51.3\%} (Retail) and \textbf{40.0\%} (Airline). \textsc{Fission-GRPO} also performs particularly well in the \textit{Base} and \textit{Miss Param} categories (e.g., 57.50\% on 8B Base and 30.50\% on 4B Miss Param), indicating more accurate function and parameter handling. Compared with specialized 8B agents, \textsc{Fission-GRPO} (Qwen3-8B) surpasses ToolACE-2-8B and BitAgent-8B on BFCL by 9.75 and 9.00 points, respectively, and by a much larger margin on TAU-Bench Retail.

\begin{figure}[t]
  \centering
  \includegraphics[width=\linewidth]{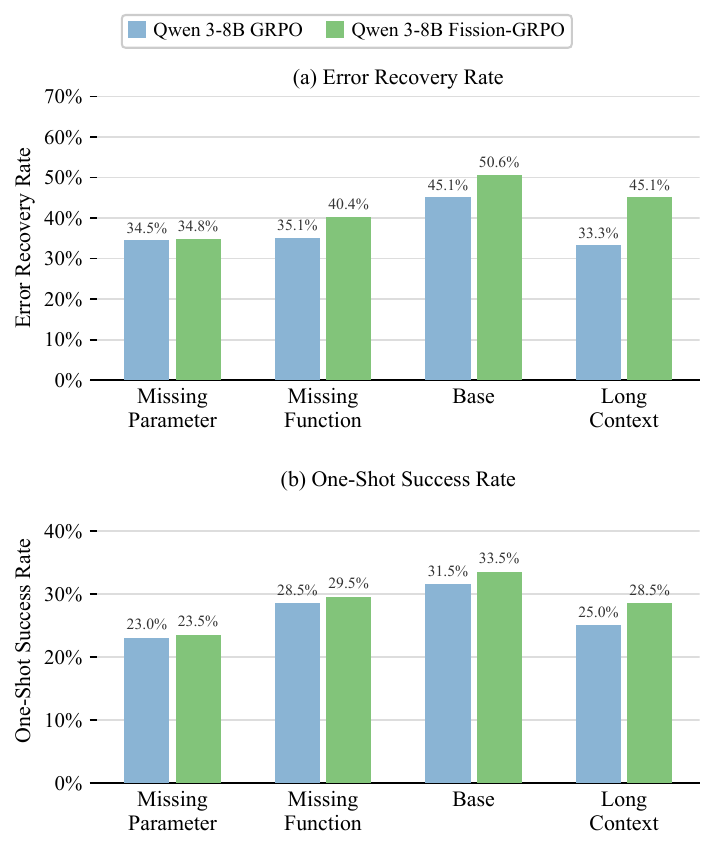}
  \caption{Performance decomposition on BFCL v4 Multi-Turn (Qwen3-8B).}
  \label{fig:recovery_breakdown}
\end{figure}

\subsection{Error Recovery Analysis}
\label{sec:error_recovery_analysis}

To identify the source of performance gains, we decompose the overall success rate into two components: \textit{One-Shot Success Rate} (success without triggering errors) and \textit{Error Recovery Rate} (success conditioned on the occurrence of an error).

Figure~\ref{fig:recovery_breakdown} illustrates this breakdown for the Qwen3-8B model. The results clearly indicate that the performance improvement is primarily driven by enhanced error recovery capabilities. \textsc{Fission-GRPO} yields an average improvement of \textbf{5.7\%} in Error Recovery Rate across all categories, with particularly substantial gains in \textit{Long Context} (+11.8\%) and \textit{Base} (+5.5\%) scenarios.

Crucially, this gain does not come at the expense of fundamental capabilities. The One-Shot Success Rate is preserved and even modestly improved by an average of 1.75\%, confirming that the fission mechanism provides complementary benefits to both error prevention and error correction.

\subsection{Impact of Feedback Quality}
\label{sec:ablation_feedback}

To disentangle the contribution of the \textit{Fission} mechanism from the informational gain of the Error Simulator, we conduct an ablation study across three settings:
(1) \textbf{GRPO}: The standard baseline without explicit recovery training.
(2) \textbf{Fission-Static}: Applies the fission update but uses a fixed, generic error message for all errors.\footnote{The static prompt is: ``ERROR: Function call failed. Please verify your output format, function name, required parameters, and parameter values are correct.''}
(3) \textbf{Fission-Dynamic}: Our full method using the Error Simulator for context-aware feedback.

\begin{table}[t]
\centering
\small
\setlength{\tabcolsep}{3.5pt}
\renewcommand{\arraystretch}{1.15}
\begin{tabular}{@{}l ccccc@{}}
\toprule
\textbf{Method} & \textbf{Avg.} & \textbf{Base} & \textbf{M.Func} & \textbf{M.Param} & \textbf{Long} \\
\midrule

\rowcolor{gray!12}
\multicolumn{6}{l}{\textit{\textbf{Qwen3-1.7B}}} \\
\midrule
GRPO             & 17.12          & 22.00          & 18.50          & 15.50          & 12.50 \\
Static           & 17.75          & 23.50          & \textbf{21.00} & 15.50          & 11.00 \\
\textbf{Dynamic} & \textbf{20.38} & \textbf{29.00} & 18.50          & \textbf{16.00} & \textbf{18.00} \\
\midrule

\rowcolor{gray!12}
\multicolumn{6}{l}{\textit{\textbf{Qwen3-4B}}} \\
\midrule
GRPO             & 36.38          & 46.50          & 34.50          & 27.50          & 37.00 \\
Static           & 37.25          & 50.50          & 34.00          & 28.50          & 36.00 \\
\textbf{Dynamic} & \textbf{40.87} & \textbf{51.50} & \textbf{42.50} & \textbf{30.50} & \textbf{39.00} \\
\midrule

\rowcolor{gray!12}
\multicolumn{6}{l}{\textit{\textbf{Qwen3-8B}}} \\
\midrule
GRPO             & 42.75          & 50.50          & 41.00          & 36.00          & 43.50 \\
Static           & 44.00          & 53.50          & 43.00          & 35.50          & 44.00 \\
\textbf{Dynamic} & \textbf{46.75} & \textbf{57.50} & \textbf{43.50} & \textbf{38.00} & \textbf{48.00} \\
\bottomrule
\end{tabular}
\caption{\textbf{Ablation on feedback quality.} \textit{Static} denotes Fission training with generic error prompts; \textit{Dynamic} uses our simulated feedback. \textit{Avg.} denotes overall accuracy; \textit{M.Func} and \textit{M.Param} denote missing function and missing parameter errors, respectively. The best result in each panel is in \textbf{bold}.}
\label{tab:ablation_feedback}
\end{table}



Results in Table~\ref{tab:ablation_feedback} show a clear progression from GRPO to Fission-Static to Fission-Dynamic. 
First, Fission-Static consistently outperforms GRPO. Even with uninformative feedback, resampling recovery attempts from failed contexts encourages the model to refine its internal state tracking (e.g., +1.25\% on Qwen3-8B Overall Acc), validating that the \textit{structural} intervention of the fission mechanism is inherently valuable.
Since Fission-Static uses a fixed generic prompt containing zero teacher-derived information, its gains over GRPO indicate that the fission mechanism itself contributes meaningful learning benefits, rather than merely inheriting information from larger models.
Second, Fission-Dynamic yields substantial additional gains, especially at 4B and 8B. The gap between Static and Dynamic (e.g., +3.62 points on Qwen3-4B) underscores the necessity of precise supervision. Generic signals fail to guide the model through complex errors, whereas simulated feedback more effectively guides learning toward correcting specific semantic errors, particularly in the \textit{Miss Param} and \textit{Long Context} subsets.

\subsection{Error Simulator Analysis}
\label{sec:simulator_analysis}

A key concern is whether the Error Simulator generalizes beyond its training domains and whether its outputs inadvertently leak ground-truth information. We evaluate both aspects through automated and human assessments across two test sets: (i) an \textbf{in-domain} set of 200 held-out error trajectories from the original training domains (excluded from simulator training), and (ii) an \textbf{out-of-domain} set of 200 real rollout failures collected from live inference in two new domains---e-commerce order management and calendar scheduling---with no additional fine-tuning of the simulator. We evaluate along three dimensions using Claude Sonnet 4 as an LLM judge (1--5 scale): \textit{Localization} (does the feedback pinpoint the fault?), \textit{Actionability} (can the student attempt a correction?), and \textit{Non-leakage} (does it avoid exposing ground-truth details?).

\begin{table}[t]
\centering
\small
\setlength{\tabcolsep}{5pt}
\renewcommand{\arraystretch}{1.1}
\begin{tabular}{@{}l c c c@{}}
\toprule
\textbf{Dimension} & \textbf{In-domain} & \textbf{Out-of-domain} & \textbf{$\Delta$} \\
\midrule
Localization   & 4.21 & 3.97 & $-$0.24 \\
Actionability  & 4.26 & 4.12 & $-$0.14 \\
Non-leakage    & 4.88 & 4.83 & $-$0.05 \\
\bottomrule
\end{tabular}
\caption{\textbf{Error Simulator cross-domain evaluation} (1--5 scale). The simulator maintains high quality on unseen domains with minimal degradation.}
\label{tab:simulator_eval}
\end{table}

As shown in Table~\ref{tab:simulator_eval}, the cross-domain drop is at most 0.24 points. Non-leakage remains near-ceiling in both in-domain and out-of-domain settings ($4.88 \to 4.83$), suggesting that the simulator does not degenerate into reproducing ground-truth calls when faced with unfamiliar schemas. To further validate this metric, two independent annotators rated 100 randomly sampled outputs, of which 96\% were judged strictly non-leaking, with Cohen's $\kappa = 0.71$ (substantial agreement), broadly consistent with the automated scores.

\subsection{Trigger Frequency and Compute Efficiency}
\label{sec:ablation_trigger_freq}

\begin{figure}[t]
  \centering
  \includegraphics[width=\linewidth]{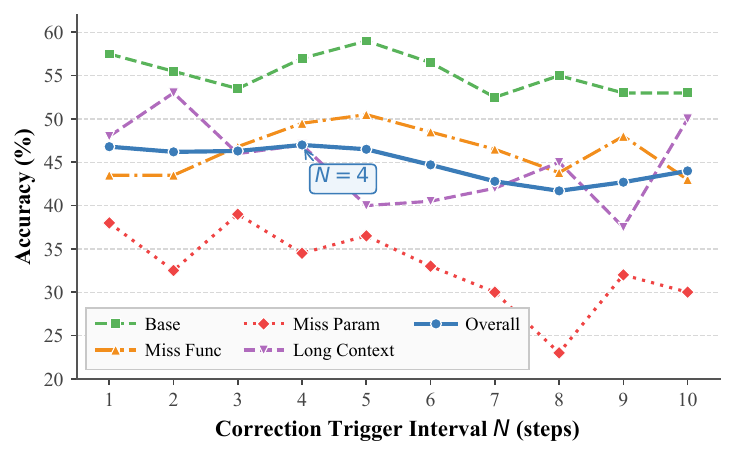}
  \caption{Multi-turn performance across different correction trigger intervals ($N$) on BFCL v4 Multi-Turn.}
  \label{fig:correction_interval}
\end{figure}

We study how the minimum trigger interval $N$ (in global steps) affects performance. $N$ limits correction updates to occur at most once every $N$ global steps; we fix the training budget to 234 standard updates and vary only $N$, using a LIFO strategy to prioritize the most recent errors.
Figure~\ref{fig:correction_interval} shows that performance remains stable for small-to-moderate $N$ but degrades noticeably as corrections become sparse. The drop is most pronounced on Miss Param and Long Context, indicating that parameter-level errors and long-context interactions particularly benefit from timely correction. This stability over a range of $N$ values highlights a practical trade-off: correction does not need to be extremely frequent to remain effective, but overly sparse schedules allow error patterns to accumulate unchecked.

\paragraph{Compute-Matched Comparison.}
To verify that fission gains reflect more effective budget allocation rather than additional compute, we set $G' = G$ so that each fission update has the same cost as a standard GRPO update, then compare both methods at the same total number of update steps under two trigger settings.

\begin{table}[t]
\centering
\small
\setlength{\tabcolsep}{3.5pt}
\renewcommand{\arraystretch}{1.1}
\begin{tabular}{@{}l l ccccc@{}}
\toprule
$N$ & \textbf{Method} & \textbf{Avg.} & \textbf{Base} & \textbf{M.F} & \textbf{M.P} & \textbf{Long} \\
\midrule
\multirow{2}{*}{3}
& GRPO (432 upd.)    & 41.75 & 56.50 & 40.00 & 29.50 & 41.00 \\
& Fission (432 upd.) & \textbf{46.75} & \textbf{57.50} & \textbf{43.50} & \textbf{38.00} & \textbf{48.00} \\
\midrule
\multirow{2}{*}{6}
& GRPO (312 upd.)    & 43.75 & \textbf{53.00} & 44.00 & 31.50 & 46.50 \\
& Fission (312 upd.) & \textbf{46.12} & \textbf{53.00} & \textbf{48.50} & \textbf{36.00} & \textbf{47.00} \\
\bottomrule
\end{tabular}
\caption{\textbf{Compute-matched comparison} on BFCL v4 Multi-Turn (Qwen3-8B). Both methods use identical total update steps.}
\label{tab:compute_matched}
\end{table}

As shown in Table~\ref{tab:compute_matched}, \textsc{Fission-GRPO} consistently outperforms GRPO under matched compute, with gains concentrated on Miss Param (+8.50 / +4.50) and Miss Func (+3.50 / +4.50)---precisely the error-recovery-heavy subsets. This confirms that fission reallocates the training budget toward the policy's failure modes rather than simply inflating compute. Representative training curves for GRPO and \textsc{Fission-GRPO} are provided in Appendix~\ref{app:training_curves}; both exhibit stable reward dynamics with no evidence of collapse or divergence.

\subsection{Case Study: Error Recovery Behaviors}
\label{sec:case_study}

To qualitatively illustrate the robustness improvements, we compare three Qwen3-8B variants (Base, GRPO, \textsc{Fission-GRPO}) on a representative multi-turn file manipulation task (from BFCL v4 Multi-Turn Base) requiring state tracking across directory changes and file moves (full logs in Appendix~\ref{app:case_study}).

We observe three distinct error-recovery patterns. 
The \textbf{Base} model exhibits \textit{collapse}: it fails to update internal state after partial command success, entering repetitive invalid retries until conversation breakdown. 
\textbf{GRPO} shows \textit{hallucination}: it recognizes errors but lacks grounding—when a file path becomes invalid, it invents non-existent parameters (e.g., a \texttt{path} argument for \texttt{ls}) rather than verifying the actual state. 
In contrast, \textbf{\textsc{Fission-GRPO}} demonstrates \textit{active diagnosis}: it employs a diagnose-then-correct strategy, deploying verification tools (e.g., \texttt{find}) to resolve state uncertainty before reattempting the task. 
This comparison shows that \textsc{Fission-GRPO} transforms error signals into active diagnostic capabilities rather than brittle heuristics.

\section{Conclusion}
\label{sec:conclusion}

We presented \textsc{Fission-GRPO}, a framework that transforms execution errors into on-policy corrective supervision for multi-turn tool use. By intercepting failures, augmenting them with simulated feedback, and resampling recovery attempts, our approach enables smaller models to learn robust self-correction rather than collapsing into repetitive loops. On BFCL v4 Multi-Turn, \textsc{Fission-GRPO} improves Qwen3-8B by 5.7\% in error recovery and 4.0\% in overall accuracy, while also showing consistent gains on TAU-Bench and TAU2-Bench. More broadly, the fission paradigm may extend to other iterative refinement domains such as code debugging and mathematical reasoning.

\section*{Limitations}

Our work has several limitations that suggest directions for future research.

\paragraph{Evaluation Scope.}
We evaluate \textsc{Fission-GRPO} on the BFCL v4 Multi-Turn benchmark and TAU-Bench / TAU2-Bench, both of which feature interactive error feedback mechanisms with retry attempts. While these benchmarks cover diverse tool APIs and error dynamics, our evaluation remains within the domain of tool-calling agents. Extending to other settings with error-retry dynamics (e.g., interactive code debugging or web navigation with fallback) is a promising direction for future work.

\paragraph{Computational Overhead.}
The fission mechanism introduces additional computational cost by resampling $G'$ rollouts for each intercepted error. Our compute-matched experiments (\S\ref{sec:ablation_trigger_freq}) show that gains persist under identical update budgets, and a configurable trigger interval $N$ allows trading off correction frequency against training efficiency. Nonetheless, scaling to very large models or environments with expensive real API calls could amplify absolute costs, and further optimization of fission scheduling remains an open direction.

\section*{Acknowledgements}
We thank the anonymous reviewers for their constructive feedback. This work is partially supported by Hong Kong RGC GRF No.\ 14206324.

\bibliography{custom}

\appendix

\section{Prompt Template for the Error Simulator}
\label{app:simulator_prompt}

\begin{figure}[!b]
  \centering
  \includegraphics[width=1\linewidth]{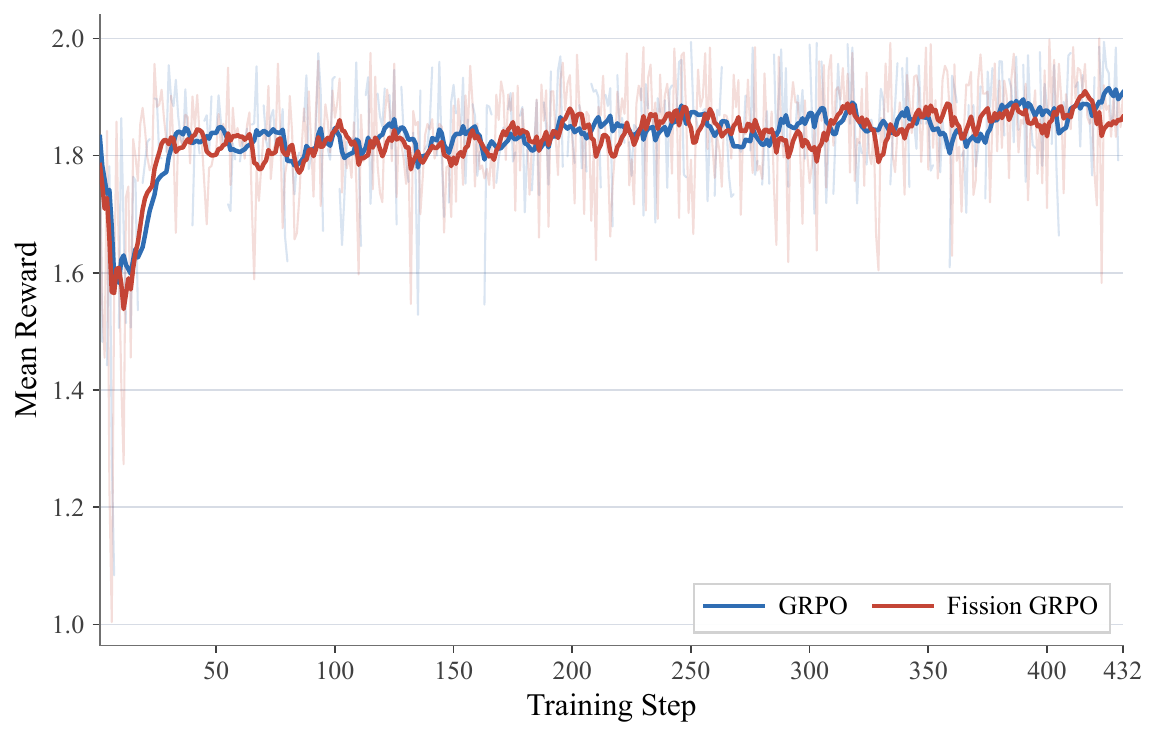}
  \caption{\textbf{Representative training reward curves for GRPO and \textsc{Fission-GRPO}.} Although step-level rewards are noisy, the smoothed trajectories remain stable throughout training, with rapid early improvement and no evidence of collapse or divergence.}
  \label{fig:training_curves}
\end{figure}

To improve reproducibility, we provide the prompting template used to query the error simulator $S_{\phi}$.
We use a two-message chat format: a system prompt that specifies the simulator role and output constraints,
followed by a user prompt that injects the original context, ground-truth tool calls, and the model's failed attempt.

\begin{figure*}[t]
  \centering

  \begin{PromptCard}{Prompt Templates}

    \captionsetup{type=lstlisting}
    \captionof{lstlisting}{System prompt for querying the error simulator.}
    \label{lst:sim_sys}
    \vspace{-0.35em}
\begin{lstlisting}
You are a Runtime Environment Simulator for an AI Agent.
Your role is to act as the API Server or Operating System that executes tool calls.

IMPORTANT CONTEXT:
You are receiving a tool call from an Agent that has ALREADY FAILED validation or logic checks against the Ground Truth.
Your task is NOT to judge correctness. Your task is to generate the specific ERROR MESSAGE that the system would return to the Agent.

GOAL:
Generate a short, realistic, and actionable error message (starting with "ERROR: ")
that will help the Agent understand why its call failed compared to the expected Ground Truth.

PRIORITY ERROR CATEGORIES:
- Dependency & Sequence Violations
- Parameter Hallucination
- Schema & Parameter Errors
- Business Logic Errors

EVALUATION LOGIC:
- Reference the Ground Truth: ground_truth is usually correct and serves as the primary standard.
- Verify Context: cross-check the Agent output against the User Request in the Original Context.
- Ambiguity Rule: if the Agent output differs from ground_truth but is still plausible, note missing validation.

OUTPUT RULES:
- Start with "ERROR: " (case-sensitive)
- Be specific: mention actual parameter names/values from the failed attempt
- Sound like a system/API response
- Keep it concise (1--2 sentences)
- Return ONLY the error string (no JSON, no markdown, no extra explanation)

ERROR MESSAGE EXAMPLES (Real error style):
<<ERROR_EXAMPLES_SNIPPET>>

CRITICAL REMINDERS:
- Do NOT output JSON like {"error": "..."}; output plain text only
- Do NOT add any preamble; only the "ERROR: ..." line
- Do NOT fabricate placeholders unless they appear in the failed attempt
- The error should be what the runtime system returns, not an analysis
\end{lstlisting}

    \vspace{0.9em}

    \captionsetup{type=lstlisting}
    \captionof{lstlisting}{User prompt template (simulation input).}
    \label{lst:sim_user}
    \vspace{-0.35em}
\begin{lstlisting}
## Simulation Task
The Agent attempted to execute a tool call, but it was INCORRECT compared to the Ground Truth.
Generate the system error message triggered by the Agent's specific mistake.

1) Original Context (what the Agent saw)

[System instructions & tools]
<<SYSTEM_AND_TOOLS>>

[Dialogue history before this attempt (non-system turns)]
<<DIALOGUE_HISTORY>>

2) Execution Comparison

[Ground-truth tool call(s)]
<<GROUND_TRUTH_TOOL_CALLS>>

[Failed tool call(s) extracted from the model output]
<<FAILED_TOOL_CALLS>>

3) Instruction

Compare the failed attempt against the ground truth under the given context.
Identify the first critical failure and generate the runtime error.

Output:
Return ONLY one error string starting with "ERROR:".
\end{lstlisting}

  \end{PromptCard}

  \caption{Two-message prompting format used to query the error simulator $S_{\phi}$.}
  \label{fig:sim_prompt}
\end{figure*}

\section{Representative Training Curves}
\label{app:training_curves}
To complement the main results, we provide representative training reward curves for GRPO and \textsc{Fission-GRPO} with Qwen3-8B in Figure~\ref{fig:training_curves}. Although step-level rewards are noisy, the smoothed trajectories exhibit similar macro-level dynamics: both methods improve rapidly during early training and remain stable thereafter, with no evidence of reward collapse or divergence. These curves further suggest that the gains of \textsc{Fission-GRPO} are achieved without introducing optimization instability.

\section{Training Algorithm Details}
\label{app:algorithm}
Algorithm~\ref{alg:fission_grpo} outlines the detailed execution flow of the \textsc{Fission-GRPO} framework. The process alternates between standard exploration (to maintain general capability and mine errors) and fission-based updates (to learn specific recovery strategies).

\begin{algorithm}[h]
\caption{Detailed Training Procedure of \textsc{Fission-GRPO}}
\label{alg:fission_grpo}
\small
\begin{algorithmic}[1]
\Require Policy $\pi_\theta$, Reference Policy $\pi_{\text{ref}}$, Error Simulator $\mathcal{S}_\phi$
\Require Training dataset $\mathcal{D}$
\Require Hyperparameters: Learning rate $\eta$, KL coefficient $\beta$, Clip ratio $\epsilon$
\Require Group sizes: $G$ (Exploration), $G'$ (Fission/Correction)
\Require Thresholds: Buffer trigger $B_{\text{trig}}$, Success score $R_{\text{thresh}}=1.0$

\State Initialize Corrective Sample Pool $\mathcal{B} \leftarrow \emptyset$ \Comment{Implemented as LIFO Stack}

\For{iteration $k=1, \dots, K$}
    \State \textcolor{blue}{\textsc{// Stage 1: Standard Exploration \& Mining}}
    \State Sample batch of user queries $x \sim \mathcal{D}$
    \State Generate exploration group $\{\tau_i\}_{i=1}^G \sim \pi_\theta(\cdot|x)$
    \State Compute rewards for each trajectory: $r_i \leftarrow R(\tau_i, t)$ \Comment{Eq.~\ref{eq:reward}, simplified in pseudocode}
    
    \State \textit{Compute GRPO Advantages:}
    \State \quad $\mu_R \leftarrow \frac{1}{G}\sum r_i, \quad \sigma_R \leftarrow \text{Std}(r_i)$
    \State \quad $\hat{A}_i \leftarrow \frac{r_i - \mu_R}{\sigma_R + \epsilon}$
    
    \State \textit{Update Policy (Standard):}
    \State \quad $\mathcal{L}_{\text{GRPO}} \leftarrow \frac{1}{G} \sum_{i=1}^G \left[ \min(\rho_i \hat{A}_i, \text{clip}(\rho_i, 1\pm\epsilon)\hat{A}_i) - \beta \mathbb{D}_{\text{KL}}(\pi_\theta \| \pi_{\text{ref}}) \right]$
    \State \quad $\theta \leftarrow \theta + \eta \nabla_\theta \mathcal{L}_{\text{GRPO}}$
    
    \State \textcolor{blue}{\textsc{// Stage 2: Synthesis \& Accumulation}}
    \State Identify error set $\mathcal{E} = \{ \tau_i \mid R_{\text{corr}}(\tau_i) < R_{\text{thresh}} \lor R_{\text{fmt}}(\tau_i) = 0 \}$
    \For{each error trajectory $\tau_{\text{err}} \in \mathcal{E}$}
        \If{$R_{\text{fmt}}(\tau_{\text{err}}) == 0$}
            \State $f \leftarrow \text{GetFormatError}(\tau_{\text{err}})$
        \Else
            \State $f \leftarrow \mathcal{S}_\phi(x, \tau_{\text{err}}, \tau_{\text{gt}})$ \Comment{Generate diagnostic feedback}
        \EndIf
        
        \State Construct corrective context $x_{\text{corr}} \leftarrow [x; \tau_{\text{err}}; f]$
        \State Compute Deduplication Key $k \leftarrow \text{Hash}(x, \tau_{\text{err}})$
        \If{$k \notin \text{Keys}(\mathcal{B})$}
            \State $\text{Push}(x_{\text{corr}}) \to \mathcal{B}$ \Comment{LIFO Push}
        \EndIf
    \EndFor
    
    \State \textcolor{blue}{\textsc{// Stage 3: Fission-Based Remedial Update}}
    \If{$|\mathcal{B}| \ge B_{\text{trig}}$}
        \State $X_{\text{batch}} \leftarrow \text{Pop}(B_{\text{trig}})$ items from top of $\mathcal{B}$ \Comment{LIFO: Fetch freshest errors}
        \State Initialize batch loss $\mathcal{L}_{\text{total}} \leftarrow 0$
        
        \For{each corrective context $x_{\text{corr}} \in X_{\text{batch}}$}
            \State \textit{Fission Resampling:}
            \State \quad Generate recovery group $\{\tau'_j\}_{j=1}^{G'} \sim \pi_\theta(\cdot|x_{\text{corr}})$
            \State \quad Compute rewards $\{r'_j\}$ for recovery attempts
            
            \State \textit{Compute Corrective Advantages:}
            \State \quad $\mu'_R \leftarrow \frac{1}{G'}\sum r'_j, \quad \sigma'_R \leftarrow \text{Std}(r'_j)$
            \State \quad $\hat{A}'_j \leftarrow \frac{r'_j - \mu'_R}{\sigma'_R + \epsilon}$ \Comment{Variance restored via Fission}
            
            \State \textit{Accumulate Gradients:}
            \State \quad $\mathcal{L}_{\text{corr}} \leftarrow \frac{1}{G'} \sum_{j=1}^{G'} \left[ \min(\rho'_j \hat{A}'_j, \dots) - \beta \mathbb{D}_{\text{KL}} \right]$
            \State \quad $\mathcal{L}_{\text{total}} \leftarrow \mathcal{L}_{\text{total}} + \mathcal{L}_{\text{corr}}$
        \EndFor
        
        \State $\theta \leftarrow \theta + \eta \nabla_\theta \mathcal{L}_{\text{total}}$ \Comment{Apply corrective update}
    \EndIf
\EndFor
\end{algorithmic}
\end{algorithm}

\section{Extended Case Study Analysis}
\label{app:case_study}

\begin{figure*}[t]
    \centering
    \includegraphics[width=\textwidth]{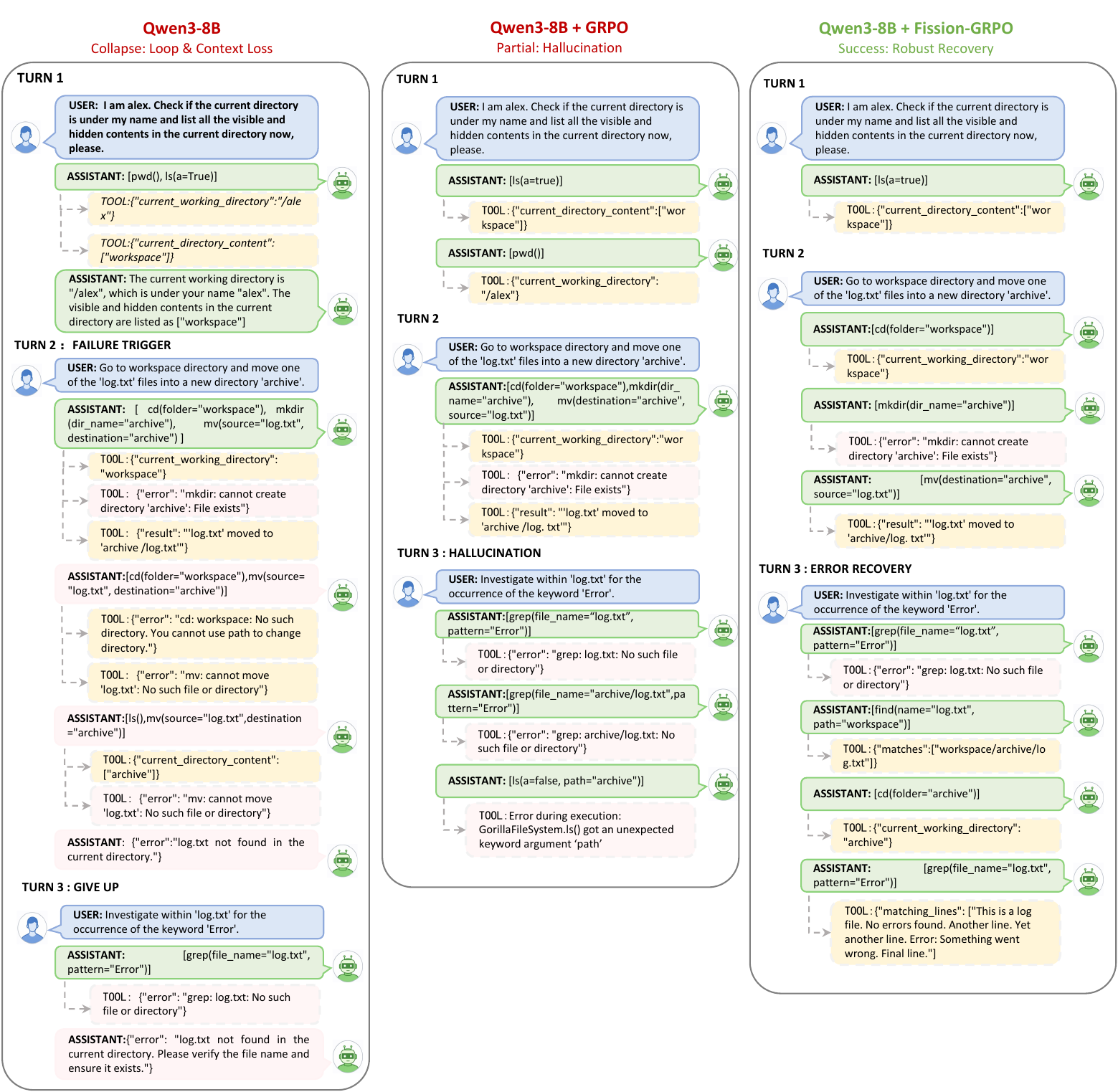}
    \caption{\textbf{Detailed visualization of Multi-turn Error Recovery.} Comparisons of trajectories generated by Qwen3-8B under different training regimes. The Base model collapses due to immediate state loss; the GRPO model suffers from latent state mismatch leading to hallucination in later turns; \textsc{Fission-GRPO} overcomes this by employing diagnostic tools (\texttt{find}) to actively resolve state uncertainties.}
    \label{fig:case_study_full}
\end{figure*}

In this section, we provide a detailed breakdown of the case study referenced in Section~\ref{sec:case_study}. Figure~\ref{fig:case_study_full} visualizes the trajectories of Qwen3-8B under three training conditions on a multi-turn file manipulation task (Sample ID: \texttt{multi\_turn\_base\_1}).

\paragraph{Scenario Overview.}
The user requests to verify the current directory, move a \texttt{log.txt} file into a new \texttt{archive} folder, and then search for a keyword within that file. The key challenge arises in Turn 2: \texttt{mkdir archive} fails (directory already exists), but \texttt{cd workspace} and \texttt{mv log.txt} succeed. This partial-success state requires careful tracking—in Turn 3, since the file was moved to \texttt{archive}, a direct \texttt{grep} will fail, requiring the agent to locate the file first.

\paragraph{Qwen3-8B (Base) --- State Awareness Collapse.}
The Base model correctly issues the initial batch command \texttt{[cd, mkdir, mv]}. However, it fails to update its internal state to reflect that it is already inside \texttt{workspace} after the successful \texttt{cd}. When attempting to handle the \texttt{mkdir} error, it redundantly retries \texttt{cd workspace}, which fails (``No such directory'' within the current directory). Confused by this feedback, it spirals into a loop of invalid operations, ultimately failing to realize the file was already moved.

\paragraph{Qwen3-8B + GRPO --- Latent State Mismatch \& Hallucination.}
The GRPO model succeeds in Turn 2 (the file is moved), but fails to track the consequence—specifically, that \texttt{log.txt} is no longer in the current directory but in the \texttt{archive} subdirectory. This latent state mismatch surfaces in Turn 3: it first tries \texttt{grep("log.txt")} (fails), then attempts a heuristic guess \texttt{grep("archive/log.txt")} (also fails). Lacking a grounded fallback strategy, it resorts to hallucination, inventing a non-existent \texttt{path} parameter for \texttt{ls}.

\paragraph{Qwen3-8B + \textsc{Fission-GRPO} --- Active Diagnosis.}
Our model handles the Turn 2 state transition correctly. More importantly, in Turn 3, when faced with the same ``No such file'' error, it demonstrates a superior recovery mechanism: instead of guessing, it deploys \texttt{find(name="log.txt", path="workspace")} to empirically verify the file's location. Using the confirmed path, it performs a precise state update via \texttt{cd(folder="archive")}, then executes \texttt{grep} successfully. This confirms that \textsc{Fission-GRPO} learns to bridge state gaps through active diagnosis rather than relying on fragile internal memory or hallucinated corrections.

\end{document}